\documentclass[runningheads]{llncs}
\usepackage[T1]{fontenc}
\usepackage{float}
\usepackage{pifont}
\usepackage{footnote}
\usepackage{enumitem}
\usepackage{bm}
\usepackage{arydshln}
\usepackage{booktabs}
\usepackage{multicol}
\usepackage{multirow}
\usepackage{color}
\usepackage{xcolor}     
\usepackage{colortbl}
\usepackage{soul}
\usepackage{bbding}
\usepackage{makecell}
\usepackage{mathtools}
\usepackage{imakeidx}
\usepackage{amssymb}
\usepackage{graphicx}
\usepackage{amsmath}
\usepackage{threeparttable}
\definecolor{citecolor}{HTML}{0071BC}
\definecolor{linkcolor}{HTML}{ED1C24}
\usepackage[colorlinks,
            anchorcolor=red,
            citecolor=citecolor, 
            linkcolor=linkcolor,
            ]{hyperref}
\makeindex
\usepackage{arydshln}
\usepackage{lipsum}
\usepackage[toc]{multitoc}
\usepackage[edges]{forest}
\usepackage[normalem]{ulem}

\usepackage{bbding}
\usepackage[most]{tcolorbox}

\usepackage{algorithm}
\usepackage{algorithmic}

\usepackage{minitoc}
\usepackage[toc,page,header]{appendix}

\definecolor{orchid}{rgb}{0.85, 0.44, 0.84}
\definecolor{rubinred}{rgb}{0.82, 0.0, 0.28}
\definecolor{flagship}{rgb}{0, 255, 0}
\definecolor{radiologist}{rgb}{0.50, 0.50, 1}

\definecolor{YT}{HTML}{002FA7}

\newcommand{\dataset}{Merlin Plus}

\newcolumntype{P}[1]{>{\centering\arraybackslash}p{#1}}
\newlength\savewidth

\usepackage{graphicx,verbatim}
\begin{document}
\title{Merlin Plus: A Large-Scale, Multi-Cancer, Image-Mask-Report Dataset}
\titlerunning{\dataset}
%
\author{
Pedro R. A. S. Bassi\inst{1,2,3}\textsuperscript{\dag} \and
Wenxuan Li\inst{1,2,3}\textsuperscript{\dag} \and
Szymon Płotka\inst{4}\textsuperscript{\dag} \and\\
Ruby Honjol\inst{5} \and
Jakub Prządo\inst{6} \and
Xinze Zhou\inst{1} \and
Kang Wang\inst{7} \and
Yang Yang\inst{7} \and\\
Malte Jensen\inst{5} \and
Akshay S. Chaudhari\inst{5} \and
Curtis P. Langlotz\inst{5} \and
Alan L. Yuille\inst{1} \and
Zongwei Zhou\inst{1,8}\thanks{Correspondence to: Zongwei Zhou (\href{mailto:zzhou82@jh.edu}{\texttt{zzhou82@jh.edu}})}
}
\authorrunning{P. Bassi et al.}
\institute{
Johns Hopkins University \and
Harvard Medical School \and
Massachusetts General Hospital \and
Jagiellonian University \and
Stanford University \and
Warmian-Masurian Cancer Center \and
University of California, San Francisco \and
Johns Hopkins Medicine
}


\maketitle              
\vspace{-0.5cm}
\begin{center}
\textsuperscript{\dag}\textit{Equal contribution}
\end{center}
\begin{abstract}

Multi-cancer segmentation in computed tomography (CT) is fundamentally limited by the scarcity of tumor masks across different organs. We present Merlin Plus, the first large-scale CT dataset with radiologist-created tumor masks across 9 organs. Merlin Plus extends the Merlin dataset by adding 1,153 per-voxel tumor masks and longitudinal metadata. To create these tumor masks, we developed a report-based active-learning framework in which radiology reports identify tumor cases for annotation and support training of a tumor segmentation model. The model generates initial masks, which radiologists review and correct to produce the final masks, reducing annotation burden while maintaining high-quality annotations. Besides tumor masks, the longitudinal metadata in Merlin Plus enables temporal modeling of cancer progression. By directly addressing the major bottleneck of limited multi-cancer segmentation masks, Merlin Plus supports scalable multi-organ cancer detection, segmentation, and longitudinal analysis in CT. Dataset is available at: \href{https://github.com/MrGiovanni/MerlinPlus}{https://github.com/MrGiovanni/MerlinPlus}

\end{abstract}

\section{Introduction}

\setcounter{footnote}{0}

Early cancer detection is critical: 5-year survival often exceeds 80\% when tumors are found early but can drop below 20\% when tumors are advanced \cite{crosby2022early}. Computed Tomography (CT) offers a major opportunity for early and incidental cancer detection. Since over 300 million CT scans are performed worldwide each year, many scans inevitably contain early-stage tumors, regardless of why these scans were requested. Because early tumors are difficult to detect, artificial intelligence (AI) models are being developed to assist radiologists, and some already surpass human performance \cite{li2026early,cao2023large}. State-of-the-art AI models for early tumor detection are often based on segmentation \cite{li2026early,cao2023large,bassi2025learning,bassi2025radgpt,plotka2026mamba,liu2023clip}, due to its superior accuracy\footnote{Segmentation models are trained with more precise supervision (per-voxel masks). Therefore, segmentation models surpass report generation models \cite{bassi2025radgpt}, CLIP-based models \cite{bassi2025learning}, and classifiers (Table \ref{tab:all_results}) in tumor detection metrics (e.g., sensitivity and specificity), even when trained on the same data.} and interpretability. Segmentation models outline tumors, allowing radiologists to better understand, verify, and trust the AI predictions. However, public segmentation models can accurately segment tumors in only a few organs. The reason is that segmentation models are trained with tumor masks (outlines of tumors drawn by radiologists), and public datasets only offer masks for tumors in a few organs. Public CT datasets mainly have masks for tumors in the liver, kidney, pancreas, colon, and lungs \cite{bassi2025radgpt,bilic2023liver,heller2021state,li2025pants,antonelli2022medical,li2025expectation,setio2017validation}. Cancers in other organs account for around half of the global cancer mortality \cite{filho2025globocan}, and remain largely unaddressed in the current AI research.

To address this gap, we present \textbf{Merlin Plus}, the first public CT dataset with masks for tumors in the spleen, bladder, gallbladder, stomach, duodenum, and prostate\footnote{The ULS and FLARE datasets \cite{de2025uls23,FLARE23-ma2024automaticorganpancancersegmentation} may include tumor masks in these organs, but they do not identify the organs where tumors are. Models trained on ULS and FLARE could not accurately detect or segment tumors in the 9 organs we evaluated (Table \ref{tab:all_results}), suggesting the datasets lack sufficient masks for these tumor types.}. Additionally, it is the largest collection of public masks for tumors in the adrenal glands, esophagus, and uterus (surpassing AbdomenAtlas 2.0 \cite{chen2025scaling}). The aim of Merlin Plus is to enable the research community to train segmentation models for these 9 tumor types, which cannot be accurately segmented or even detected by public AI models today (Table \ref{tab:all_results}). The original Merlin dataset provided 25,494 CT-Report pairs \cite{blankemeier2026merlin}, acquired at the Stanford Hospital. In Merlin Plus, we have created 1,153 public tumor masks for these scans, including all CT scans in Merlin with malignant tumors in these 9 organs (according to reports), and 325 benign tumors. Additionally, Merlin Plus provides comprehensive metadata, including age, sex, race, CT scanner types, de-identified patient IDs, and scan dates, which enable audits of demographic and protocol bias~\cite{chen2026benchx}. This new longitudinal metadata allows training AI models to analyze cancer progression over time. 

To create Merlin Plus, segmentation models generated initial tumor masks, which one of four radiologists reviewed and revised to create the final tumor masks. We used a report-based active-learning pipeline based on the Report Supervision (R-Super \cite{bassi2025learning,bassi2025scaling}) segmentation model. First, R-Super was trained directly from radiology reports, without tumor masks. Learning from reports is important because masks for these 9 tumor types are scarce or absent in public CT datasets, and public segmentation models cannot segment them accurately. R-Super then generated masks for radiologist revision and was retrained with the revised masks and reports in successive active-learning cycles. Reports were also used to automatically correct minor mask errors and identify inaccurate masks: masks that most disagreed with the reported tumor description were prioritized for radiologist revision. This workflow made tumor mask creation 3 to 5$\times$ faster than manual annotation alone.

To demonstrate the value of Merlin Plus, we trained state-of-the-art tumor segmentation models on it, and compared them to public tumor segmentation and detection models. We demonstrate that segmentation models trained on the CT scans, reports and masks from Merlin Plus significantly surpass public AI models in the detection and segmentation of the 9 tumor types evaluated here (Table \ref{tab:all_results}). Merlin Plus segmentation models also significantly surpassed models trained on the original Merlin dataset (Table \ref{tab:all_results}). Overall, our main contributions are as follows:
\begin{enumerate}
    \item \textbf{Merlin Plus dataset}: we create public tumor segmentation masks for 9 tumor types with few or no masks in public CT datasets. Merlin Plus enables multi-tumor detection and segmentation substantially better than public AI models and models trained on Merlin (Table \ref{tab:all_results}).
    \item \textbf{Clinical Metadata}: Merlin Plus provides key metadata absent from Merlin, such as de-identified patient IDs and CT dates, patient demographics (age, sex, race) and scan details (e.g., scanner manufacturer, kVp values).
    \item \textbf{Report-based active learning}: our new active-learning strategy is more effective for developing multi-tumor segmentation datasets and allows radiologists to create tumor segmentation masks 3 to 5$\times$ faster.
\end{enumerate}

\section{Merlin Plus Dataset}

Merlin Plus includes the 25,494 CT scans and corresponding radiology reports from the Merlin dataset \cite{blankemeier2026merlin}. Merlin is a public dataset collected from the Stanford hospital PACS, it comprises CT examinations acquired between 2012 and 2018. The studies were identified through the Stanford Medicine Research Data Repository (STARR) using CPT codes 72192, 72193, 72194, 74150, 74160, 74170, 74176, 74177, and 74178.

Merlin Plus adds three key data elements to Merlin: per-voxel tumor segmentation masks, patient IDs, and CT scan dates. The tumor masks enable precise tumor segmentation and improve AI-based tumor detection. Patient IDs and scan dates identify scans from the same patient and define the interval between scans, enabling longitudinal analyses and the development of longitudinal AI algorithms. Of a total of 18,317 patients, 3,830 have 2 or more scans, and 1,524 have 3 or more. We provide a total of 1,153 tumor masks, for tumors in 9 organs: 186 spleen tumor masks, 122 bladder, 96 gallbladder, 54 esophagus, 160 stomach, 78 duodenum, 115 uterus, 80 prostate, 262 adrenal glands. All masks were manually verified and corrected (if needed) by radiologists. These masks focus on tumor types with few or no public tumor masks on CT. These tumor masks include all scans where the radiology report indicated malignancy in these 9 organs, plus 325 benign tumors. CT scans are mainly abdominal, but often include the pelvis and chest.

\begin{figure}[h]
    \includegraphics[width=\textwidth]{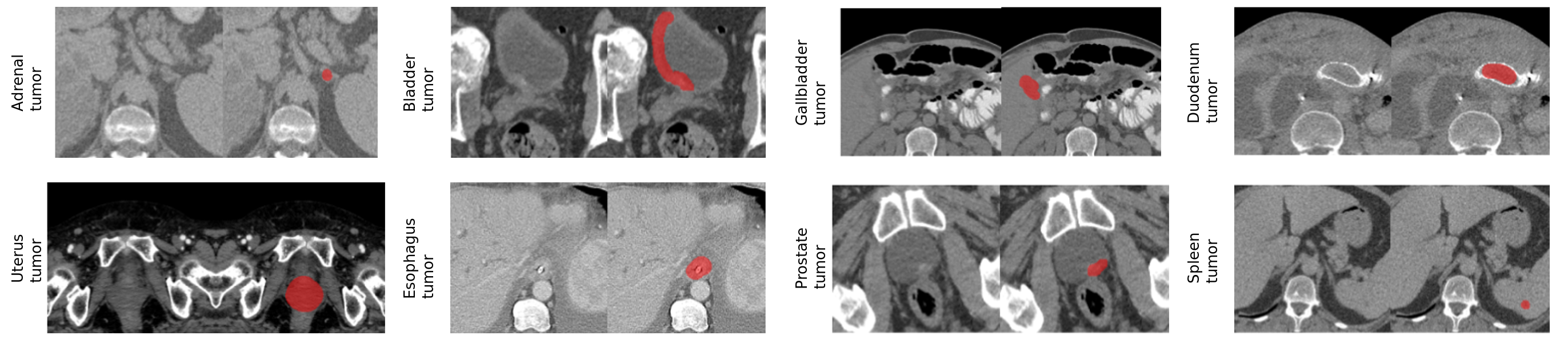}
    \caption{\textbf{Examples of different tumor masks provided in \textit{Merlin Plus}.} Masks are 3D, and were created by 4 radiologists, assisted by report-based active learning.}
    \label{fig:dataset_description}
\end{figure}

\section{Report-based Active Learning}
\label{sec:reportbasedactivelearning}

\begin{figure}[t]
    \centering
    \includegraphics[width=1\linewidth]{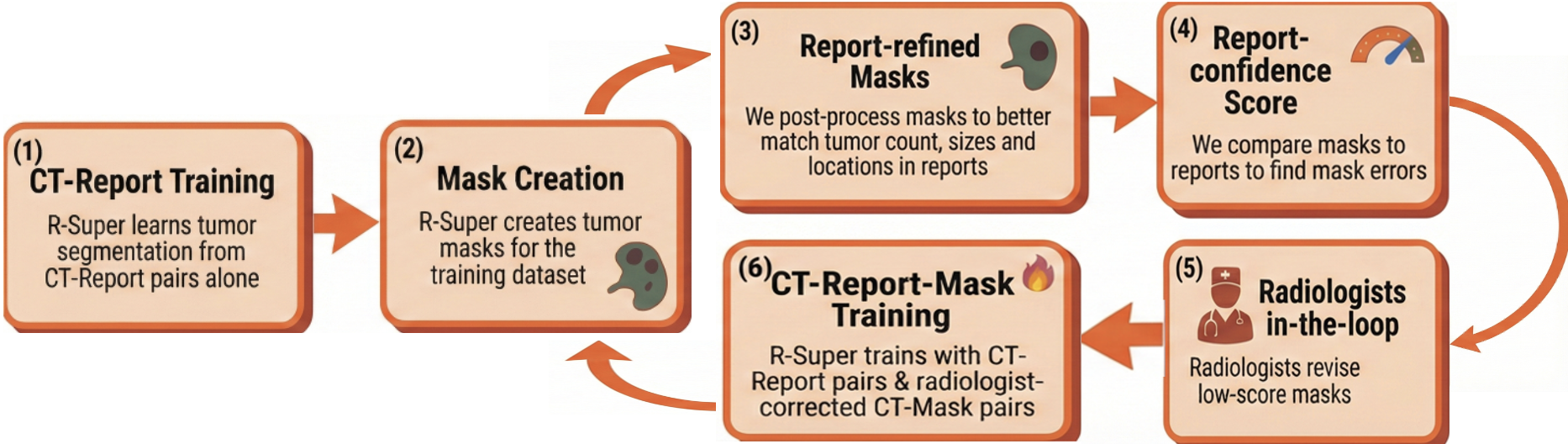}
    \caption{\textbf{Report-based Active Learning.} Our active learning strategy has 6 steps. Steps 2-6 are repeated cyclically. (1) we train a tumor segmentation model from radiology reports (no mask); (2) the segmentation model generates tumor segmentation masks; (3) we automatically refine these masks to be more consistent with the tumor descriptions in radiology reports; (4) and we calculate a report-confidence score, which flags mask errors by comparing masks and reports; (5) masks with the lowest scores are revised by radiologists; and (6) we fine-tune the segmentation model using all available CT-Report pairs and CT-Mask pairs.}
    \label{fig:active_learning}
\end{figure}

The Merlin Plus multi-tumor masks were created by 4 radiologists. To enable our team to create 1,153 multi-tumor masks in just 5 months, we developed an active learning strategy centered on radiology reports (Fig. \ref{fig:active_learning}). Our framework addresses the cold-start problem: the requirement for a pre-existing high-quality segmentation model to begin the active learning loop~\cite{zhou2017fine,zhou2021active,chen2023making}. Correcting low-quality masks from a poor initial model often takes as long as drawing them from scratch. Since no public segmenter accurately covers our 9 CT tumor types, we initialize our loop using the Report Supervision (R-Super) \cite{bassi2025learning,bassi2025scaling} training framework to train a segmentation model without tumor masks. This report-supervised initialization allows our radiologists to begin the cycle by correcting reasonable AI-made tumor masks rather than drawing from scratch~\cite{zhou2025efficient}. All final masks were verified by radiologists.

The report-based active learning process (Fig. \ref{fig:active_learning}) accelerated annotation by 3 to 5$\times$. We measured annotation time on 20 randomly selected scans. The same radiologist annotated each case manually and by revising the R-Super-made mask. To reduce human learning effects, the order of the assisted and unassisted conditions was alternated across cases and reversed in the second round. For scans with only small tumors ($\leq$2 cm), annotation averaged 3 minutes with AI-assistance versus 16 minutes without; for larger tumors (>2 cm), it averaged 18 versus 52 minutes. The active learning cycle is described below. In total, we used 3 cycles, creating around 450 radiologist-revised masks in each cycle.

\noindent
\textbf{Step 1: learning multi-tumor segmentation from radiology reports.}
We trained R-Super to segment tumors, learning only from CT scans and radiology reports (no mask). R-Super teaches segmentation models to segment tumors consistent with report-described tumor location, number, and size.

\noindent
\textbf{Step 2: mask creation.}
We ran inference with R-Super on its training set and saved the resulting soft tumor masks (per-voxel tumor probabilities).

\noindent
\textbf{Step 3: report-refined masks.}
We post-process the soft tumor masks to make them match the tumor descriptions in radiology reports as closely as possible. This procedure is based on the Ball Loss of R-Super \cite{bassi2025learning}. It automatically fixes some common mask errors (e.g., tumors segmented in wrong organs), reducing the amount of corrections radiologists will need to perform in Step 4.

\textit{(A) report attribute extraction.} Before active learning, an LLM extracts from the reports the tumor \emph{count}, \emph{diameters}, and \emph{locations} (tumor organ and, when available, tumor slice). We use a zero-shot LLM (Llama 3.1 70B AWQ), previously shown to achieve 96\% accuracy for tumor information extraction \cite{bassi2025radgpt}. Each report is processed once and the extracted information is saved.

\textit{(B) erase tumor probabilities outside the region where the tumor is reported}. We keep probabilities only inside the reported tumor organ. Organs are located with pre-saved organ masks, created by an nnU-Net \cite{isensee2021nnu} trained on AbdomenAtlas 1.1 \cite{li2024well,li2024abdomenatlas}. Organ masks are dilated by 2\,cm to compensate for organ segmentation inaccuracies (often small \cite{bassi2024touchstone}) and for tumors that grow beyond the organ. If a tumor slice (tumor z coordinate) is provided, we also erase tumor probabilities in slices farther than one tumor diameter from the reported slice.

\textit{(C) enforce the segmented tumor number and diameters to match the report.} First, we locate the most probable location of each tumor mentioned in the report. To this end, we use Ball Convolutions \cite{bassi2025learning}. These are standard 3D convolutions with a non-learnable kernel containing a binary ball, whose diameter matches the tumor diameter in the report. The convolution moves this ball across the soft tumor mask. At each kernel position the convolution sums all tumor probabilities inside the ball. Thus, the convolution output is maximum where the ball is over the most probable location for a tumor matching the ball diameter, the highest probability ball. We use sequential Ball Convolutions to locate each tumor in the report, from largest to smallest, ignoring probabilities in previously selected highest-probability balls to avoid locating the same tumor twice. After locating all tumors, we set the tumor probabilities outside all highest-probability balls to zero, enforcing the segmented tumor number to match the report, and their diameters to not surpass the report.

\textit{(D) convert the soft tumor probabilities into a binary tumor mask, enforcing segmented tumors to match the tumor \emph{sizes} in the report.} For each highest probability ball, we estimate the segmented tumor volume by counting voxels with probability $\geq 0.5$. We also estimate the corresponding reported tumor volume from the reported tumor diameters, using a sphere volume approximation if the report provides one tumor diameter, and an ellipsoid volume approximation if 2 or 3 diameters are provided (if 2 are provided, we estimate the third diameter as the average of the other 2). If the segmented and reported tumor volumes agree within 30\%, we just threshold the tumor probabilities at 0.5 inside the highest probability ball. Otherwise, we refine the mask to match the report volume: we sort the voxels inside the ball by descending tumor probability, and set the most probable voxels to 1 until the reported volume is reached; remaining voxels are set to 0. The final binary masks are named report-refined masks.


\noindent
\textbf{Step 4: report-confidence score.}
Step 3 enforces that masks match the report description, but this does not guarantee that the report-refined mask captures the true tumor. In Step 4, we identify report-refined masks that are likely unreliable and need radiologist review~\cite{chen2026large}. Our key intuition is that the amount of correction in Step 3 is informative: if the segmentation model confidently segments tumors that closely match the report descriptions, Step 3 makes minor corrections to the soft mask. Otherwise, Step 3 must substantially correct the mask, and the refined mask may satisfy the report description but still be inaccurate (e.g., tumor at an incorrect location within the reported organ). These cases will be flagged by a low \emph{report-confidence score}. We define this score as the Dice coefficient between the soft tumor mask (Step 2) and the binary report-refined mask (Step 3). High scores indicate few corrections and higher reliability; low scores indicate many corrections (e.g., inserting missed tumors, binarizing low confidence tumors). For example, if the report describes a spleen tumor but the model misses it, Step 3 will place a tumor within the spleen anyway (wherever tumor probabilities are higher), but the soft mask will have a low Dice score with this tumor, flagging the case as unreliable.

\noindent
\textbf{Step 5: radiologists correct unreliable masks.} Radiologists analyze and correct the report-refined masks with the lowest report-confidence score (e.g., the 50 masks with the lowest score per tumor type). Radiologists were provided with the report-refined masks and the radiology report.

\noindent
\textbf{Step 6: fine-tuning with reports and masks.} We fine-tune R-Super using all CT-Report pairs plus all corrected CT-Mask pairs accumulated up to the current active learning cycle~\cite{zhang2024leveraging}. This teaches R-Super to avoid the errors it previously made. By training with masks and reports, we obtain a more accurate segmentation than training with masks alone (Table \ref{tab:all_results}), improving active learning efficiency.

\section{Results}

\begin{table}[!tp]
\centering
\setlength{\tabcolsep}{1.4pt}
\renewcommand{\arraystretch}{1.04}
\caption{\textbf{The tumor masks in Merlin Plus improve tumor detection and segmentation:} models trained in Merlin Plus surpass models trained in Merlin (no tumor mask) and public models. Table shows tumor detection (Se, Sp, F1) and segmentation (DSC); top: internal, bottom: external validation. Positive scans per organ (test masks in parentheses). Internal: spleen 113 (29), bladder 43 (18), gallbladder 53 (19), adrenal 157 (16), esophagus 20 (10), stomach 57 (35), duodenum 36 (11), uterus 81 (21), prostate 31 (17) (591 total; 114 controls, 176 masks). External: spleen 357 (31), bladder 146 (32), gallbladder 174 (10), adrenal 357 (39), esophagus 189 (23), stomach 624 (71), duodenum 74 (-), uterus 131 (17), prostate 35 (-) (2087 total; 178 controls, 223 masks). \textbf{Voxtell} (adrenal): internal 48/94/63/20, external 50/81/63/5 (Se/Sp/F1/DSC). Grey inter-reader row: two-radiologist DSC agreement (90 masks). Detection metrics 95\% CI: bootstrap (1000); DSC: 1.96\,std$/\sqrt{n}$.}

\sbox0{%
\tiny
\begin{tabular}{@{}l*{2}{c}*{20}{c}@{}}
\toprule
\multicolumn{23}{c}{\scriptsize \textbf{internal validation (Merlin Plus)}} \\
\midrule
\tiny model & \multicolumn{2}{c}{\scriptsize supervision} & \multicolumn{4}{c}{\scriptsize spleen} & \multicolumn{4}{c}{\scriptsize bladder} & \multicolumn{4}{c}{\scriptsize gallbladder} & \multicolumn{4}{c}{\scriptsize adrenal} & \multicolumn{4}{c}{\scriptsize esophagus} \\
\cmidrule(lr){2-3}\cmidrule(lr){4-7}\cmidrule(lr){8-11}\cmidrule(lr){12-15}\cmidrule(lr){16-19}\cmidrule(lr){20-23}
& report & mask & Se & Sp & F1 & DSC & Se & Sp & F1 & DSC & Se & Sp & F1 & DSC & Se & Sp & F1 & DSC & Se & Sp & F1 & DSC \\
\midrule
\multicolumn{23}{l}{\textit{public multi-tumor segmentation model}} \\
ULS \cite{de2025uls23} &  & x & 34 & 78 & 44 & 3 & 42 & 66 & 35 & 12 & 23 & 92 & 33 & 3 & 2 & 100 & 4 & 0 & 17 & 99 & 25 & 0 \\
FLARE25 baseline \cite{FLARE23-ma2024automaticorganpancancersegmentation} &  & x & 12 & 100 & 21 & 4 & 12 & 100 & 21 & 0 & 13 & 100 & 23 & 1 & 17 & 100 & 28 & 6 & 0 & 100 & 0 & 0 \\
FLARE23 top1 \cite{FLARE23-ma2024automaticorganpancancersegmentation} &  & x & 20 & 99 & 32 & 3 & 2 & 100 & 5 & 0 & 21 & 100 & 34 & 13 & 13 & 100 & 24 & 5 & 0 & 100 & 0 & 0 \\
FLARE24 top1 \cite{FLARE23-ma2024automaticorganpancancersegmentation} &  & x & 19 & 100 & 31 & 2 & 5 & 100 & 9 & 0 & 19 & 100 & 32 & 9 & 20 & 100 & 33 & 11 & 0 & 100 & 0 & 0 \\
\midrule
\multicolumn{23}{l}{\textit{trained on \textbf{Merlin}}} \\
classification & x &  & 58 & 58 & 58 & - & 46 & 80 & 46 & - & 72 & 38 & 47 & - & 57 & 67 & 63 & - & 55 & 83 & 44 & - \\
\midrule
\multicolumn{23}{l}{\textit{trained on \textbf{Merlin Plus}}} \\
MedFormer~\cite{gao2022data} &  & x & 66 & 84 & 72 & 9 & 79 & 85 & 72 & 21 & 76 & 72 & 64 & 20 & 76 & 85 & 82 & 26 & 75 & 66 & 40 & 25 \\
R-Super \cite{bassi2025learning} & x & x & 86 & 77 & \textbf{82} & \textbf{25} & 86 & 91 & \textbf{82} & \textbf{29} & 77 & 93 & \textbf{80} & \textbf{43} & 83 & 96 & \textbf{89} & \textbf{41} & 80 & 71 & \textbf{46} & \textbf{33} \\
\textcolor{gray}{\quad 95\% CI ($\pm$ half-w.)} &  &  & \textcolor{gray}{6} & \textcolor{gray}{8} & \textcolor{gray}{5} & \textcolor{gray}{9} & \textcolor{gray}{11} & \textcolor{gray}{5} & \textcolor{gray}{9} & \textcolor{gray}{10} & \textcolor{gray}{11} & \textcolor{gray}{5} & \textcolor{gray}{8} & \textcolor{gray}{13} & \textcolor{gray}{6} & \textcolor{gray}{3} & \textcolor{gray}{4} & \textcolor{gray}{15} & \textcolor{gray}{25} & \textcolor{gray}{6} & \textcolor{gray}{25} & \textcolor{gray}{17} \\
\midrule
\tiny model & \multicolumn{2}{c}{\scriptsize supervision} & \multicolumn{4}{c}{\scriptsize stomach} & \multicolumn{4}{c}{\scriptsize duodenum} & \multicolumn{4}{c}{\scriptsize uterus} & \multicolumn{4}{c}{\scriptsize prostate} & \multicolumn{4}{c}{\scriptsize average} \\
\cmidrule(lr){2-3}\cmidrule(lr){4-7}\cmidrule(lr){8-11}\cmidrule(lr){12-15}\cmidrule(lr){16-19}\cmidrule(lr){20-23}
\midrule
ULS \cite{de2025uls23} &  & x & 47 & 59 & 41 & 4 & 28 & 93 & 37 & 2 & 49 & 68 & 51 & 13 & 60 & 79 & 50 & 27 & 34 & 82 & 35 & 7 \\
FLARE25 baseline \cite{FLARE23-ma2024automaticorganpancancersegmentation} &  & x & 30 & 97 & 44 & 2 & 11 & 96 & 19 & 0 & 50 & 87 & 59 & 21 & 32 & 87 & 36 & 8 & 20 & 96 & 28 & 5 \\
FLARE23 top1 \cite{FLARE23-ma2024automaticorganpancancersegmentation} &  & x & 26 & 100 & 42 & 1 & 6 & 97 & 10 & 0 & 5 & 100 & 10 & 3 & 3 & 100 & 6 & 2 & 11 & 100 & 18 & 3 \\
FLARE24 top1 \cite{FLARE23-ma2024automaticorganpancancersegmentation} &  & x & 30 & 98 & 45 & 3 & 11 & 99 & 20 & 0 & 26 & 99 & 40 & 18 & 6 & 99 & 12 & 2 & 15 & 100 & 25 & 5 \\
\midrule
classification & x &  & 61 & 52 & 48 & - & 56 & 68 & 44 & - & 67 & 68 & 63 & - & 52 & 80 & 46 & - & 58 & 66 & 51 & - \\
\midrule
MedFormer~\cite{gao2022data} &  & x & 75 & 87 & 75 & 3 & 66 & 68 & 49 & 2 & 92 & 61 & 73 & 25 & 74 & 95 & 77 & 32 & 75 & 78 & 67 & 18 \\
R-Super \cite{bassi2025learning} & x & x & 77 & 90 & \textbf{78} & \textbf{16} & 78 & 70 & \textbf{57} & \textbf{9} & 73 & 87 & \textbf{76} & \textbf{47} & 84 & 92 & \textbf{79} & \textbf{51} & 80 & 85 & \textbf{74} & \textbf{33} \\
\textcolor{gray}{\quad 95\% CI ($\pm$ half-w.)} &  &  & \textcolor{gray}{12} & \textcolor{gray}{4} & \textcolor{gray}{9} & \textcolor{gray}{7} & \textcolor{gray}{14} & \textcolor{gray}{8} & \textcolor{gray}{12} & \textcolor{gray}{8} & \textcolor{gray}{10} & \textcolor{gray}{6} & \textcolor{gray}{7} & \textcolor{gray}{14} & \textcolor{gray}{14} & \textcolor{gray}{4} & \textcolor{gray}{12} & \textcolor{gray}{18} & \textcolor{gray}{4} & \textcolor{gray}{2} & \textcolor{gray}{4} & \textcolor{gray}{4} \\
\midrule[\heavyrulewidth]
\multicolumn{23}{c}{\scriptsize \textbf{external validation}} \\
\midrule
\tiny model & \multicolumn{2}{c}{\scriptsize supervision} & \multicolumn{4}{c}{\scriptsize spleen} & \multicolumn{4}{c}{\scriptsize bladder} & \multicolumn{4}{c}{\scriptsize gallbladder} & \multicolumn{4}{c}{\scriptsize adrenal} & \multicolumn{4}{c}{\scriptsize esophagus} \\
\cmidrule(lr){2-3}\cmidrule(lr){4-7}\cmidrule(lr){8-11}\cmidrule(lr){12-15}\cmidrule(lr){16-19}\cmidrule(lr){20-23}
\midrule
\multicolumn{23}{l}{\textit{public multi-tumor segmentation model}} \\
ULS \cite{de2025uls23} &  & x & 66 & 51 & 69 & 4 & 62 & 72 & 63 & 9 & 47 & 82 & 56 & 0 & 15 & 96 & 26 & 0 & 43 & 78 & 52 & 11 \\
FLARE25 baseline \cite{FLARE23-ma2024automaticorganpancancersegmentation} &  & x & 20 & 96 & 32 & 12 & 7 & 99 & 13 & 4 & 3 & 98 & 7 & 6 & 1 & 100 & 2 & 1 & 6 & 98 & 12 & 2 \\
FLARE23 top1 \cite{FLARE23-ma2024automaticorganpancancersegmentation} &  & x & 28 & 97 & 43 & 9 & 1 & 99 & 1 & 0 & 3 & 98 & 6 & 3 & 6 & 100 & 11 & 1 & 0 & 100 & 0 & 0 \\
FLARE24 top1 \cite{FLARE23-ma2024automaticorganpancancersegmentation} &  & x & 22 & 95 & 36 & 13 & 1 & 100 & 1 & 0 & 3 & 99 & 7 & 5 & 1 & 100 & 3 & 1 & 7 & 100 & 14 & 4 \\
\midrule
\multicolumn{23}{l}{\textit{trained on \textbf{Merlin}}} \\
classification & x &  & 59 & 51 & 64 & - & 68 & 40 & 56 & - & 59 & 42 & 54 & - & 55 & 60 & 63 & - & 78 & 60 & 72 & - \\

\midrule
\multicolumn{23}{l}{\textit{trained on \textbf{Merlin Plus}}} \\
MedFormer~\cite{gao2022data} &  & x & 65 & 67 & 72 & 23 & 57 & 86 & 65 & 27 & 66 & 70 & 67 & 5 & 83 & 83 & 86 & 26 & 88 & 77 & \textbf{84} & 22 \\
R-Super \cite{bassi2025learning} & x & x & 74 & 78 & \textbf{80} & \textbf{53} & 75 & 79 & \textbf{75} & \textbf{32} & 71 & 72 & \textbf{71} & \textbf{19} & 88 & 92 & \textbf{92} & \textbf{50} & 84 & 82 & 83 & \textbf{32} \\
\textcolor{gray}{\quad 95\% CI ($\pm$ half-w.)} &  &  & \textcolor{gray}{4} & \textcolor{gray}{6} & \textcolor{gray}{3} & \textcolor{gray}{8} & \textcolor{gray}{7} & \textcolor{gray}{6} & \textcolor{gray}{5} & \textcolor{gray}{8} & \textcolor{gray}{7} & \textcolor{gray}{6} & \textcolor{gray}{5} & \textcolor{gray}{16} & \textcolor{gray}{3} & \textcolor{gray}{4} & \textcolor{gray}{2} & \textcolor{gray}{8} & \textcolor{gray}{6} & \textcolor{gray}{6} & \textcolor{gray}{4} & \textcolor{gray}{9} \\
\midrule
\textcolor{gray}{inter-reader} &  &  & \textcolor{gray}{-} & \textcolor{gray}{-} & \textcolor{gray}{-} & \textcolor{gray}{63} & \textcolor{gray}{-} & \textcolor{gray}{-} & \textcolor{gray}{-} & \textcolor{gray}{48} & \textcolor{gray}{-} & \textcolor{gray}{-} & \textcolor{gray}{-} & \textcolor{gray}{52} & \textcolor{gray}{-} & \textcolor{gray}{-} & \textcolor{gray}{-} & \textcolor{gray}{58} & \textcolor{gray}{-} & \textcolor{gray}{-} & \textcolor{gray}{-} & \textcolor{gray}{49} \\
\midrule
\tiny model & \multicolumn{2}{c}{\scriptsize supervision} & \multicolumn{4}{c}{\scriptsize stomach} & \multicolumn{4}{c}{\scriptsize duodenum} & \multicolumn{4}{c}{\scriptsize uterus} & \multicolumn{4}{c}{\scriptsize prostate} & \multicolumn{4}{c}{\scriptsize average} \\
\cmidrule(lr){2-3}\cmidrule(lr){4-7}\cmidrule(lr){8-11}\cmidrule(lr){12-15}\cmidrule(lr){16-19}\cmidrule(lr){20-23}
\midrule
ULS \cite{de2025uls23} &  & x & 44 & 62 & 57 & 2 & 47 & 82 & 49 & - & 77 & 74 & 73 & 12 & 83 & 75 & 53 & - & 54 & 74 & 56 & 5 \\
FLARE25 baseline \cite{FLARE23-ma2024automaticorganpancancersegmentation} &  & x & 17 & 95 & 29 & 2 & 12 & 97 & 20 & - & 48 & 94 & 61 & 13 & 17 & 96 & 24 & - & 15 & 97 & 22 & 6 \\
FLARE23 top1 \cite{FLARE23-ma2024automaticorganpancancersegmentation} &  & x & 18 & 100 & 31 & 0 & 10 & 100 & 17 & - & 9 & 99 & 17 & 1 & 3 & 99 & 5 & - & 9 & 99 & 15 & 2 \\
FLARE24 top1 \cite{FLARE23-ma2024automaticorganpancancersegmentation} &  & x & 15 & 97 & 26 & 2 & 7 & 98 & 12 & - & 35 & 97 & 50 & 10 & 0 & 97 & 0 & - & 10 & 98 & 17 & 5 \\
\midrule
classification & x &  & 44 & 62 & 57 & - & 49 & 65 & 42 & - & 80 & 70 & 73 & - & 57 & 61 & 32 & - & 61 & 57 & 57 & - \\
\midrule
MedFormer~\cite{gao2022data} &  & x & 59 & 65 & 70 & 2 & 57 & 85 & 59 & - & 83 & 77 & 78 & 22 & 89 & 95 & 83 & - & 72 & 78 & 74 & 18 \\
R-Super \cite{bassi2025learning} & x & x & 65 & 78 & \textbf{76} & \textbf{13} & 74 & 79 & \textbf{66} & - & 78 & 89 & \textbf{81} & \textbf{38} & 89 & 97 & \textbf{86} & - & 78 & 83 & \textbf{79} & \textbf{34} \\
\textcolor{gray}{\quad 95\% CI ($\pm$ half-w.)} &  &  & \textcolor{gray}{4} & \textcolor{gray}{6} & \textcolor{gray}{3} & \textcolor{gray}{4} & \textcolor{gray}{10} & \textcolor{gray}{6} & \textcolor{gray}{9} & \textcolor{gray}{-} & \textcolor{gray}{8} & \textcolor{gray}{3} & \textcolor{gray}{6} & \textcolor{gray}{14} & \textcolor{gray}{10} & \textcolor{gray}{3} & \textcolor{gray}{9} & \textcolor{gray}{-} & \textcolor{gray}{2} & \textcolor{gray}{2} & \textcolor{gray}{2} & \textcolor{gray}{4} \\
\midrule
\textcolor{gray}{inter-reader} &  &  & \textcolor{gray}{-} & \textcolor{gray}{-} & \textcolor{gray}{-} & \textcolor{gray}{46} & \textcolor{gray}{-} & \textcolor{gray}{-} & \textcolor{gray}{-} & \textcolor{gray}{47} & \textcolor{gray}{-} & \textcolor{gray}{-} & \textcolor{gray}{-} & \textcolor{gray}{58} & \textcolor{gray}{-} & \textcolor{gray}{-} & \textcolor{gray}{-} & \textcolor{gray}{54} & \textcolor{gray}{-} & \textcolor{gray}{-} & \textcolor{gray}{-} & \textcolor{gray}{53} \\
\bottomrule
\end{tabular}%
}%
\resizebox{\textwidth}{!}{\usebox0}
\label{tab:all_results}
\end{table}

To demonstrate the value of Merlin Plus to tumor detection and segmentation, we trained 2 state-of-the-art tumor segmentation models on Merlin Plus: MedFormer \cite{gao2022data}, trained with masks only, and R-Super \cite{bassi2025learning,bassi2025scaling}, trained with reports and masks. We compared these models to a classifier (MedFormer encoder with a classification head) trained on Merlin, using classification labels (tumor presence/absence, per organ) from radiology reports, extracted by the Llama 3.1 70B LLM (which has 96\% accuracy in extracting tumor labels from reports \cite{bassi2025radgpt}). All models were trained with the MedFormer standard training hyper-parameters, as they are based on the MedFormer architecture. We perform internal validation on a subset of Merlin Plus (\textit{N}=591), and external validation on a private dataset (not from Stanford, \textit{N}=1044). We also compared the Merlin Plus models to leading public tumor segmentation models: the top performing models in FLARE23 and 24, the FLARE25 baseline, and ULS. For adrenal tumors, we also compare to Voxtell \cite{rokuss2026voxtell}. Notably, ULS was trained only on crops with tumors, creating a disadvantage for tumor detection \cite{de2025uls23}.

To evaluate tumor detection, we follow \cite{bassi2025learning,bassi2025radgpt} and use an organ-level, volume-based thresholding: we consider a tumor detected if the number of segmented tumor voxels inside the organ where the report mentions tumors is above a validation-defined threshold. Tumors segmented outside the correct organ (e.g., spleen tumor in the head) are considered incorrect. Organs are located with organ segmentation masks created by a segmentation model (nnU-Net) trained on AbdomenAtlas 3.0 \cite{bassi2025radgpt}. Organ masks are dilated by 2 cm to compensate for segmentation errors. For fairness, the table shows sensitivity, specificity and F1-Score at the operating point of highest average sensitivity and specificity for all models. DSC is calculated only for cases with tumors (and tumor masks).

\textbf{The masks in \dataset\ substantially improve tumor detection and segmentation performance}. Our main contribution in \dataset\ is the creation of ground-truth tumor masks for 9 tumor types missing in current CT segmentation datasets. Table \ref{tab:all_results} shows the importance of these masks: models trained on Merlin Plus (with our masks) substantially surpassed the classifier trained on the original Merlin (no mask) in multi-tumor detection, by large margins. Also, models trained on Merlin Plus substantially surpassed the state-of-the-art public multi-tumor segmentation models by large margins (Voxtell, ULS, and FLARE). This indicates that the tumors considered here are underrepresented in previous public datasets.

\textbf{DSC and inter-reader agreement}. Inter-reader agreement was low for the tumor types studied here: the average DSC between masks created by two radiologists was 53\% (Table~\ref{tab:all_results}), compared with a median inter-reader DSC of 86\% reported for pancreatic tumors \cite{li2025pants}. This low agreement indicates that these tumors are difficult to identify and delineate on CT, explaining the relatively low DSC of all AI models. Because of this difficulty, CT is not the primary detection modality for many of these tumors. For example, transvaginal ultrasound (TVS) is usually used for detecting uterine tumors, endoscopy for esophageal tumors, and cystoscopy for bladder tumors. However, CT is performed at a far greater scale than these examinations (300M annually). AI-assisted analysis of routine CT could therefore allow incidental detection at scale: AI could help radiologists identify suspicious tumors and prompt targeted examinations for confirmation, allowing earlier cancer detection in asymptomatic patients.

\section{Conclusion} 

The lack of public segmentation masks has long limited multi-tumor detection and segmentation. We show that training on multi-tumor masks yields substantially better tumor segmentation and detection. This gain comes from the precise, per-voxel supervision that masks provide, which is more informative than classification labels. Notably, a segmentation model trained on Merlin Plus with reports and masks outperformed a classification model trained on Merlin by over +22\% in tumor detection F1-Score. Merlin Plus models also surpassed the state-of-the-art public multi-tumor segmentation models (Voxtell, FLARE, and ULS). Overall, Merlin Plus provides public resources that can accelerate multi-tumor detection and segmentation, and move AI toward the goal of leveraging the 300M CT scans performed annually for incidental cancer detection.

\begin{credits}
\subsubsection{\ackname} This work was supported by the Lustgarten Foundation for Pancreatic Cancer Research and the National Institutes of Health (NIH) under Award Number R01EB037669. This research was funded in part by National Science Center, Poland 2025/57/B/ST6/03100. We would like to thank the Johns Hopkins Research IT team in \href{https://researchit.jhu.edu/}{IT@JH} for their support and infrastructure resources where some of these analyses were conducted; especially \href{https://researchit.jhu.edu/research-hpc/}{DISCOVERY HPC}. We thank Jaimie Patterson for writing a news article about this project. Paper content is covered by patents pending.

\subsubsection{\discintname}
The authors declare no competing interests.
\end{credits}
%
%
%
\clearpage
\bibliographystyle{splncs04}
\bibliography{refs,zzhou}

\begin{thebibliography}{10}
\providecommand{\url}[1]{\texttt{#1}}
\providecommand{\urlprefix}{URL }
\providecommand{\doi}[1]{https://doi.org/#1}

\bibitem{antonelli2022medical}
Antonelli, M., Reinke, A., Bakas, S., Farahani, K., Kopp-Schneider, A., Landman, B.A., Litjens, G., Menze, B., Ronneberger, O., Summers, R.M., et~al.: The medical segmentation decathlon. Nature communications  \textbf{13}(1), ~4128 (2022)

\bibitem{bassi2025learning}
Bassi, P.R., Li, W., Chen, J., Zhu, Z., Lin, T., Decherchi, S., Cavalli, A., Wang, K., Yang, Y., Yuille, A.L., Zhou, Z.: Learning segmentation from radiology reports. In: International Conference on Medical Image Computing and Computer-Assisted Intervention. pp. 305--315. Springer (2025)

\bibitem{bassi2024touchstone}
Bassi, P.R., Li, W., Tang, Y., Isensee, F., et~al.: Touchstone benchmark: Are we on the right way for evaluating ai algorithms for medical segmentation? Advances in Neural Information Processing Systems (NeurIPS) Datasets and Benchmarks Track  \textbf{37},  15184--15201 (2024)

\bibitem{bassi2025radgpt}
Bassi, P.R., Yavuz, M.C., Hamamci, I.E., Er, S., Chen, X., Li, W., Menze, B., Decherchi, S., Cavalli, A., Wang, K., Yang, Y., Yuille, A., Zhou, Z.: Radgpt: Constructing 3d image-text tumor datasets. In: Proceedings of the IEEE/CVF International Conference on Computer Vision. pp. 23720--23730 (2025)

\bibitem{bassi2025scaling}
Bassi, P.R., Zhou, X., Li, W., P{\l}otka, S., et~al.: Scaling artificial intelligence for multi-tumor early detection with more reports, fewer masks. arXiv preprint arXiv:2510.14803  (2025)

\bibitem{bilic2023liver}
Bilic, P., Christ, P., Li, H.B., Vorontsov, E., Ben-Cohen, A., Kaissis, G., Szeskin, A., Jacobs, C., Mamani, G.E.H., Chartrand, G., et~al.: The liver tumor segmentation benchmark (lits). Medical image analysis  \textbf{84},  102680 (2023)

\bibitem{blankemeier2026merlin}
Blankemeier, L., Kumar, A., Cohen, J.P., Liu, J., Liu, L., Van~Veen, D., Gardezi, S.J.S., Yu, H., Paschali, M., Chen, Z., et~al.: Merlin: a computed tomography vision--language foundation model and dataset. Nature pp. 1--11 (2026)

\bibitem{cao2023large}
Cao, K., Xia, Y., Yao, J., Han, X., Lambert, L., Zhang, T., Tang, W., Jin, G., Jiang, H., Fang, X., et~al.: Large-scale pancreatic cancer detection via non-contrast ct and deep learning. Nature medicine  \textbf{29}(12),  3033--3043 (2023)

\bibitem{chen2023making}
Chen, L., Bai, Y., Huang, S., Lu, Y., Wen, B., Yuille, A.L., Zhou, Z.: Making your first choice: To address cold start problem in medical active learning. In: Medical Imaging with Deep Learning. pp. 496--525. PMLR (2023), \url{https://github.com/cliangyu/CSVAL}

\bibitem{chen2026benchx}
Chen, Q., Li, W., Bassi, P.R., Zhou, X., Wasserthal, J., Hamamci, I.E., Er, S., Kumar, A., Ye, Y., Wang, Y., Zhou, Y., Chaudhari, A.S., Langlotz, C., Wang, K., Yang, Y., Yuille, A.L., Zhou, Z.: {BenchX}: Benchmarking {AI} models for cancer detection and localization with demographic and protocol biases. arXiv preprint arXiv:2606.24883  (2026)

\bibitem{chen2025scaling}
Chen, Q., Zhou, X., Liu, C., Chen, H., Li, W., et~al.: Scaling tumor segmentation: Best lessons from real and synthetic data. In: Proceedings of the IEEE International Conference on Computer Vision (ICCV). pp. 24001--24013 (2025)

\bibitem{chen2026large}
Chen, Y., Zhou, Z., Li, W., Yuille, A.: Large-scale label quality assessment for medical segmentation via a vision-language judge and synthetic data. In: IEEE International Symposium on Biomedical Imaging (ISBI). pp.~1--5. IEEE (2026), \url{https://github.com/Schuture/SegAE}

\bibitem{crosby2022early}
Crosby, D., Bhatia, S., Brindle, K.M., Coussens, L.M., Dive, C., Emberton, M., Esener, S., Fitzgerald, R.C., Gambhir, S.S., Kuhn, P., et~al.: Early detection of cancer. Science  \textbf{375}(6586),  eaay9040 (2022)

\bibitem{filho2025globocan}
Filho, A.M., Laversanne, M., Ferlay, J., Colombet, M., Pi{\~n}eros, M., Znaor, A., Parkin, D.M., Soerjomataram, I., Bray, F.: The globocan 2022 cancer estimates: data sources, methods, and a snapshot of the cancer burden worldwide. International journal of cancer  \textbf{156}(7),  1336--1346 (2025)

\bibitem{gao2022data}
Gao, Y., Zhou, M., Liu, D., Yan, Z., Zhang, S., Metaxas, D.N.: A data-scalable transformer for medical image segmentation: architecture, model efficiency, and benchmark. arXiv preprint arXiv:2203.00131  (2022)

\bibitem{de2025uls23}
de~Grauw, M., Scholten, E.T., Smit, E.J., Rutten, M.J., Prokop, M., van Ginneken, B., Hering, A.: The uls23 challenge: A baseline model and benchmark dataset for 3d universal lesion segmentation in computed tomography. Medical image analysis  \textbf{102},  103525 (2025)

\bibitem{heller2021state}
Heller, N., Isensee, F., Maier-Hein, K.H., Hou, X., Xie, C., Li, F., Nan, Y., Mu, G., Lin, Z., Han, M., et~al.: The state of the art in kidney and kidney tumor segmentation in contrast-enhanced ct imaging: Results of the kits19 challenge. Medical Image Analysis  \textbf{67},  101821 (2021)

\bibitem{isensee2021nnu}
Isensee, F., Jaeger, P.F., Kohl, S.A., Petersen, J., Maier-Hein, K.H.: nnu-net: a self-configuring method for deep learning-based biomedical image segmentation. Nature Methods  \textbf{18}(2),  203--211 (2021)

\bibitem{li2026early}
Li, W., Bassi, P.R.A.S., Wu, L., Zhou, X., Zhao, Y., Chen, Q., Plotka, S., et~al.: Early and prediagnostic detection of pancreatic cancer from computed tomography. arXiv preprint arXiv:2601.22134  (2026)

\bibitem{li2025expectation}
Li, W., Bassi, P.R., Lin, T., Chou, Y.C., Wasserthal, J., Zhou, X., Chen, Q., Isensee, F., Kirchhoff, Y., Rokuss, M., et~al.: Expectation-maximization as the engine of scalable medical intelligence. arXiv preprint arXiv:2501.03410  (2025)

\bibitem{li2024abdomenatlas}
Li, W., Qu, C., Chen, X., Bassi, P.R., Shi, Y., Lai, Y., Yu, Q., Xue, H., Chen, Y., Lin, X., Tang, Y., Cao, Y., Han, H., Zhang, Z., Liu, J., Zhang, T., Ma, Y., Wang, J., Zhang, G., Yuille, A., Zhou, Z.: Abdomenatlas: A large-scale, detailed-annotated, \& multi-center dataset for efficient transfer learning and open algorithmic benchmarking. Medical Image Analysis p. 103285 (2024)

\bibitem{li2024well}
Li, W., Yuille, A., Zhou, Z.: How well do supervised {3D} models transfer to medical imaging tasks? In: International Conference on Learning Representations (2024)

\bibitem{li2025pants}
Li, W., Zhou, X., Chen, Q., Lin, T., Bassi, P.R., Chen, X., Ye, C., Zhu, Z., Ding, K., Li, H., Wang, K., Yang, Y., Tang, Y., Xu, D., Yuille, A.L., Zhou, Z.: Pants: The pancreatic tumor segmentation dataset. In: Conference on Neural Information Processing Systems (NeurIPS) Datasets and Benchmarks Track (2025)

\bibitem{liu2023clip}
Liu, J., Zhang, Y., Chen, J.N., Xiao, J., Lu, Y., Landman, B.A., Yuan, Y., Yuille, A., Tang, Y., Zhou, Z.: Clip-driven universal model for organ segmentation and tumor detection. In: Proceedings of the IEEE/CVF International Conference on Computer Vision. pp. 21152--21164 (2023)

\bibitem{FLARE23-ma2024automaticorganpancancersegmentation}
Ma, J., Zhang, Y., Gu, S., Ge, C., Wang, E., Zhou, Q., Huang, Z., Lyu, P., He, J., Wang, B.: Automatic organ and pan-cancer segmentation in abdomen ct: the flare 2023 challenge (2024), \url{https://arxiv.org/abs/2408.12534}

\bibitem{plotka2026mamba}
P{\l}otka, S., Mert, G., Chrabaszcz, M., Szczurek, E., Sitek, A.: Mamba goes home: Hierarchical soft mixture-of-experts for 3d medical image segmentation. Advances in Neural Information Processing Systems  \textbf{38},  97871--97909 (2026)

\bibitem{rokuss2026voxtell}
Rokuss, M., Langenberg, M., Kirchhoff, Y., Isensee, F., Hamm, B., Ulrich, C., Regnery, S., Bauer, L., Katsigiannopulos, E., Norajitra, T., Maier-Hein, K.: Voxtell: Free-text promptable universal 3d medical image segmentation. In: Proceedings of the IEEE/CVF Conference on Computer Vision and Pattern Recognition. pp. 37538--37557 (2026)

\bibitem{setio2017validation}
Setio, A.A.A., Traverso, A., De~Bel, T., Berens, M.S., van~den Bogaard, C., Cerello, P., Chen, H., Dou, Q., Fantacci, M.E., Geurts, B., et~al.: Validation, comparison, and combination of algorithms for automatic detection of pulmonary nodules in computed tomography images: the luna16 challenge. Medical image analysis  \textbf{42},  1--13 (2017)

\bibitem{zhang2024leveraging}
Zhang, T., Chen, X., Qu, C., Yuille, A., Zhou, Z.: Leveraging {AI} predicted and expert revised annotations in interactive segmentation: Continual tuning or full training? In: IEEE International Symposium on Biomedical Imaging (ISBI). IEEE (2024)

\bibitem{zhou2025efficient}
Zhou, X., Zhao, Y., Zhuang, C., Yu, D., Yuille, A.L., Zhou, Z.: Efficient human-in-the-loop pancreatic tumor annotation via large-scale pre-trained model with adaptive post-processing. In: IEEE International Symposium on Biomedical Imaging (ISBI). pp.~1--4. IEEE (2025), \url{https://github.com/ChrisXzz/EfficientAnno}

\bibitem{zhou2017fine}
Zhou, Z., Shin, J., Zhang, L., Gurudu, S., Gotway, M., Liang, J.: Fine-tuning convolutional neural networks for biomedical image analysis: actively and incrementally. In: IEEE/CVF Conference on Computer Vision and Pattern Recognition (CVPR). pp. 7340--7351 (2017)

\bibitem{zhou2021active}
Zhou, Z., Shin, J.Y., Gurudu, S.R., Gotway, M.B., Liang, J.: Active, continual fine tuning of convolutional neural networks for reducing annotation efforts. Medical Image Analysis  \textbf{71},  101997 (2021)

\end{thebibliography}

\end{document}